\documentclass[11pt,a4paper]{article}

\usepackage{naaclhlt2019}
\usepackage{times}
\usepackage{latexsym}
\usepackage{amsmath}
\usepackage{url}
\usepackage{graphicx}
\usepackage{hhline}
\usepackage{multirow}
\usepackage{subcaption}
\usepackage{booktabs}
\usepackage{tabularx}

\aclfinalcopy 

\newenvironment{itemizesquish}{\begin{list}{\labelitemi}{\setlength{\itemsep}{0em}\setlength{\labelwidth}{0.5em}\setlength{\leftmargin}{\labelwidth}\addtolength{\leftmargin}{\labelsep}}}{\end{list}}

\title{Combining Distant and Direct Supervision for Neural Relation Extraction}

\author{{Iz Beltagy~~~~~~Kyle Lo~~~~~~Waleed Ammar} \\
Allen Institute for Artificial Intelligence, Seattle, WA, USA\\
{\tt $\{$beltagy,kylel,waleeda$\}$@allenai.org}\\}

\date{}

\begin{document}
\maketitle

\setlength{\abovedisplayskip}{3pt}
\setlength{\belowdisplayskip}{3pt}

\begin{figure*}[ht!]
    \centering
    \includegraphics[scale=0.40]{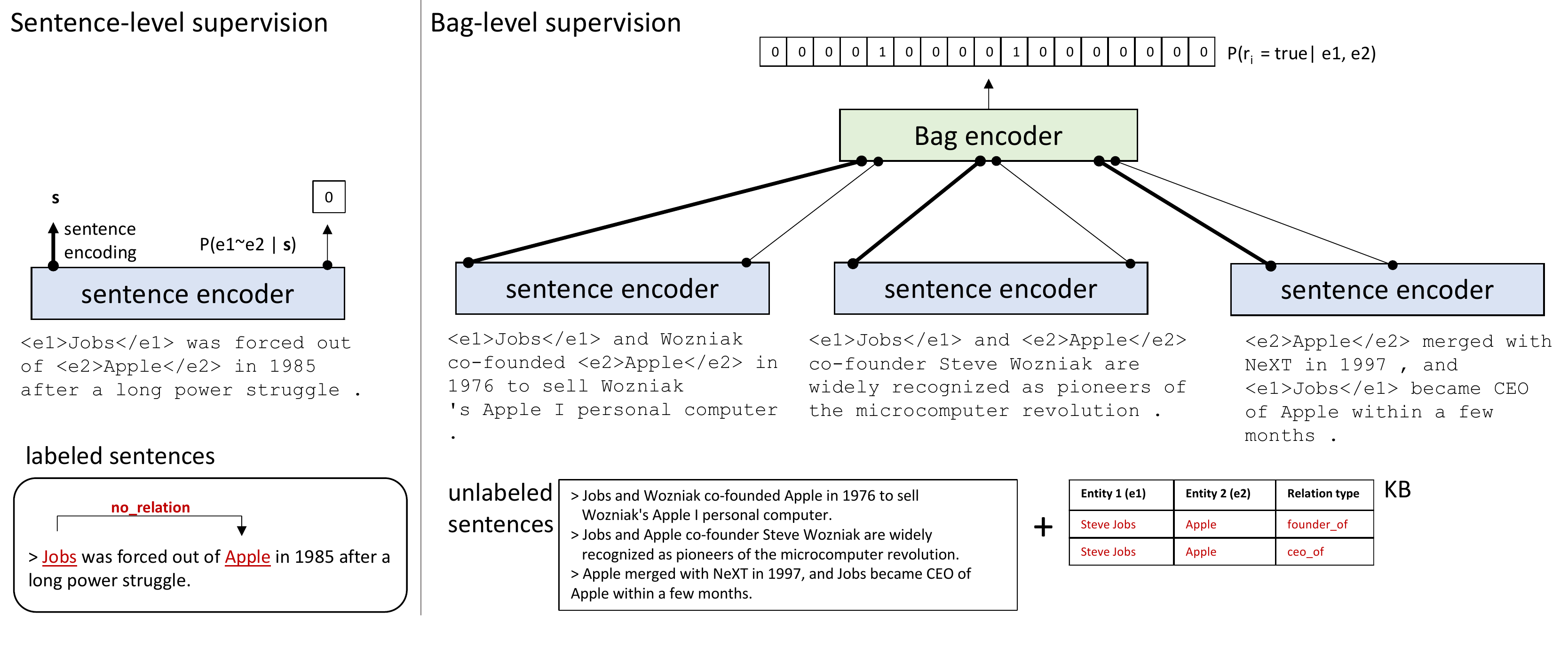}
    \vspace{-5pt}
  \caption{An overview of our approach for combining distant and  direct supervision.
   The left side shows one sentence in the labeled data and how it is used to provide direct supervision for the sentence encoder.
  The right side shows snippets of the text corpus and the knowledge base, which are then combined to construct one training instance for the model, with a bag of three input sentences and two active relations: `founder\_of' and `ceo\_of'.}
    \label{fig:relex_overview}
    \vspace{-1ex}
\end{figure*} 
\begin{abstract}
In relation extraction with distant supervision, noisy labels make it difficult to train quality models.
Previous neural models addressed this problem using an attention mechanism that attends to sentences that are likely to express the relations.
We improve such models by combining the distant supervision data with an additional directly-supervised data, which we use as supervision for the attention weights.
We find that joint training on both types of supervision leads to a better model because it improves the model's ability to identify noisy sentences.
In addition, we find that sigmoidal attention weights with max pooling achieves better performance over the commonly used weighted average attention in this setup. 
Our proposed method\footnote{\url{https://github.com/allenai/comb_dist_direct_relex/}} achieves a new state-of-the-art result on the widely used FB-NYT dataset.

\end{abstract}

\section{Introduction}

Early work in relation extraction from text used directly supervised methods, e.g., \newcite{bunescu:emnlp05}, which motivated the development of relatively small datasets with sentence-level annotations such as ACE 2004/2005, BioInfer and SemEval 2010 Task 8.
Recognizing the difficulty of annotating text with relations, especially when the number of relation types of interest is large, others~\citep{mintz:acl09,craven:ismb99} introduced 
the distant supervision approach of relation extraction, where a knowledge base (KB) and a text corpus are used to automatically generate a large dataset of labeled \emph{bags of sentences}
(a set of sentences that might express the relation) which are then used to train a relation classifier. The large number of labeled instances produced with distant supervision make it a practical alternative to manual annotations. 

However, distant supervision implicitly assumes that
all the KB facts are mentioned in the text (at least one of the sentences in each bag expresses the relation) and that 
all relevant facts are in the KB (use entities that are not related in the KB as negative examples).
These two assumptions are generally not true, which 
introduces many noisy examples in the training set.
Although many methods have been proposed to deal with such noisy training
data \cite[e.g.,][]{hoffmann:acl11,surdeanu:emnlp12,roth:cikm13,fan:acl14,zeng:emnlp15,jiang:coling16,liu:emnlp17},
a rather obvious approach has been understudied: 
combine distant supervision data with additional direct supervision.
Intuitively, directly supervising the model can improve its performance by helping it identify which of the input sentences for a given pair of entities are more likely to express a relation. 

A straightforward way to combine distant and direct supervision is to concatenate instances from both datasets into one large dataset. We show in Section~\ref{sec:ablation} that this approach doesn't help the model.  ~\citet{pershina:acl14} also observed similar results; instead, they train a graphical model on the distantly supervised instances while using the directly labeled instances to supervise a subcomponent of the model.  We discuss prior work in more detail in Section~\ref{sec:related}.

In our paper, we demonstrate a similar approach with neural networks.  Specifically, our neural model attends over sentences to distinguish between sentences that are likely to express some relation between the entities and sentences that do not.
We use the additional direct supervision to supervise these attention weights.
We train this model jointly on both types of supervision in a multitask learning setup. 
In addition, we experimentally find that sigmoidal attention weights with max pooling achieves better performance in this model setup than the commonly used weighted average attention. 





The contributions of this paper are as follows:
\begin{itemizesquish} 
\item We propose an effective neural network model for improving
distant supervision by combining it with a directly 
supervised data in the form of sentence-level annotations. The model is trained 
jointly on both types of supervision in a multitask learning setup, where the direct supervision data 
is employed as supervision for attention weights. 
\item We show experimentally that our model setup benefits from sigmoidal attention weights with max pooling over the commonly used softmax-based weighted averaging attention.
\item Our best model achieves a new state-of-the-art result
on the FB-NYT dataset, previously used by \newcite{lin:acl16,vashishth:emnlp18}. Specifically, combining both forms of supervision achieves a 4.4\% relative 
AUC increase 
than our baseline without the additional supervision.
\end{itemizesquish}
The following section defines the notation we use, describes the problem and provides an overview of our approach.

\section{Overview}
Our goal is to predict which relation types are expressed between a pair of entities ($e_1, e_2$), given all sentences in which both entities are mentioned in a large collection of unlabeled documents.

Following previous work on distant supervision, we use known tuples $(e_1, r, e_2)$ in a knowledge base $\cal{K}$ to automatically annotate sentences where both entities are mentioned with the relation type $r$.
In particular, we group all sentences $s$ with one or more mentions of an entity pair $(e_1, e_2)$ into a bag of sentences $B_{e_1,e_2}$, then automatically annotate this bag with the set of relation types $L^{\text{distant}} = \{ r \in {\cal{R}} : (e_1, r, e_2) \in {\cal{K}}\}$, where $\cal{R}$ is the set of relations we are interested in. 
We use `positive instances' to refer to cases where $|L| > 0$, and `negative instances' when $|L| = 0$.

In this paper, we leverage an existing dataset of direct supervision for relations. Each direct supervision instance consists of a token sequence $s$ containing mentions of an entity pair $(e_1, e_2)$ and one relation type (or `no relation').
We do not require that the entities or relation types in the direct supervision annotations align with those in the KB. Furthermore, we replace the relation label associated with each sentence with a binary indicator of $1$ if the sentence expresses one of the relationships of interest and $0$ otherwise.

Figure~\ref{fig:relex_overview} illustrates how we modify neural architectures commonly used in distant supervision, e.g., \newcite{lin:acl16,liu:emnlp17} to effectively incorporate
direct supervision.
The model consists of two components:
1) A \textbf{sentence encoder} (displayed in blue) reads a sequence of tokens and their relative distances from $e_1$ and $e_2$, and outputs a vector $\mathbf{s}$ representing the sentence encoding, as well as $P(e_1 \sim e_2 \mid \mathbf{s})$ representing the probability that the two entities are related given this sentence. 
2) The \textbf{bag encoder} (displayed in green) reads the encoding of each sentence in the bag for the pair $(e_1, e_2)$ and predicts $P(r = 1 \mid e_1, e_2), \forall r \in {\cal{R}}$. 

We combine both 
types of
supervision in a multi-task learning setup by minimizing the weighted sum of the cross entropy losses for $P(e_1 \sim e_2 \mid \mathbf{s})$ and $P(r = 1 \mid e_1,e_2)$.
By sharing the parameters of sentence encoders used to compute either loss, the sentence encoders become less susceptible to the noisy bag labels.
The bag encoder also benefits from the direct supervision by using the supervised distribution $P(e_1 \sim e_2 \mid \mathbf{s})$ to decide the weight of each sentence in the bag.

\section{Model}
\label{sec:model}
The model predicts a set of relation types ${L^{\text{pred}} \subset {\cal{R}}}$ given a pair of entities $e_1, e_2$ and a bag of sentences $B_{e_1, e_2}$. 
In this section, we first describe the sentence encoder part of the model (Figure~\ref{fig:relex_nn}, bottom), then describe the bag encoder (Figure~\ref{fig:relex_nn}, top), then we explain how the two types of supervision are jointly used for training the model end-to-end. 


\subsection{Sentence Encoder Architecture}
Given a sequence of words $w_1,\ldots,w_{|s|}$ in a sentence $s$, a sentence encoder translates
this sequence into a fixed length vector $\mathbf{s}$.

\begin{figure}[t!]
    \centering
    \includegraphics[scale=0.55]{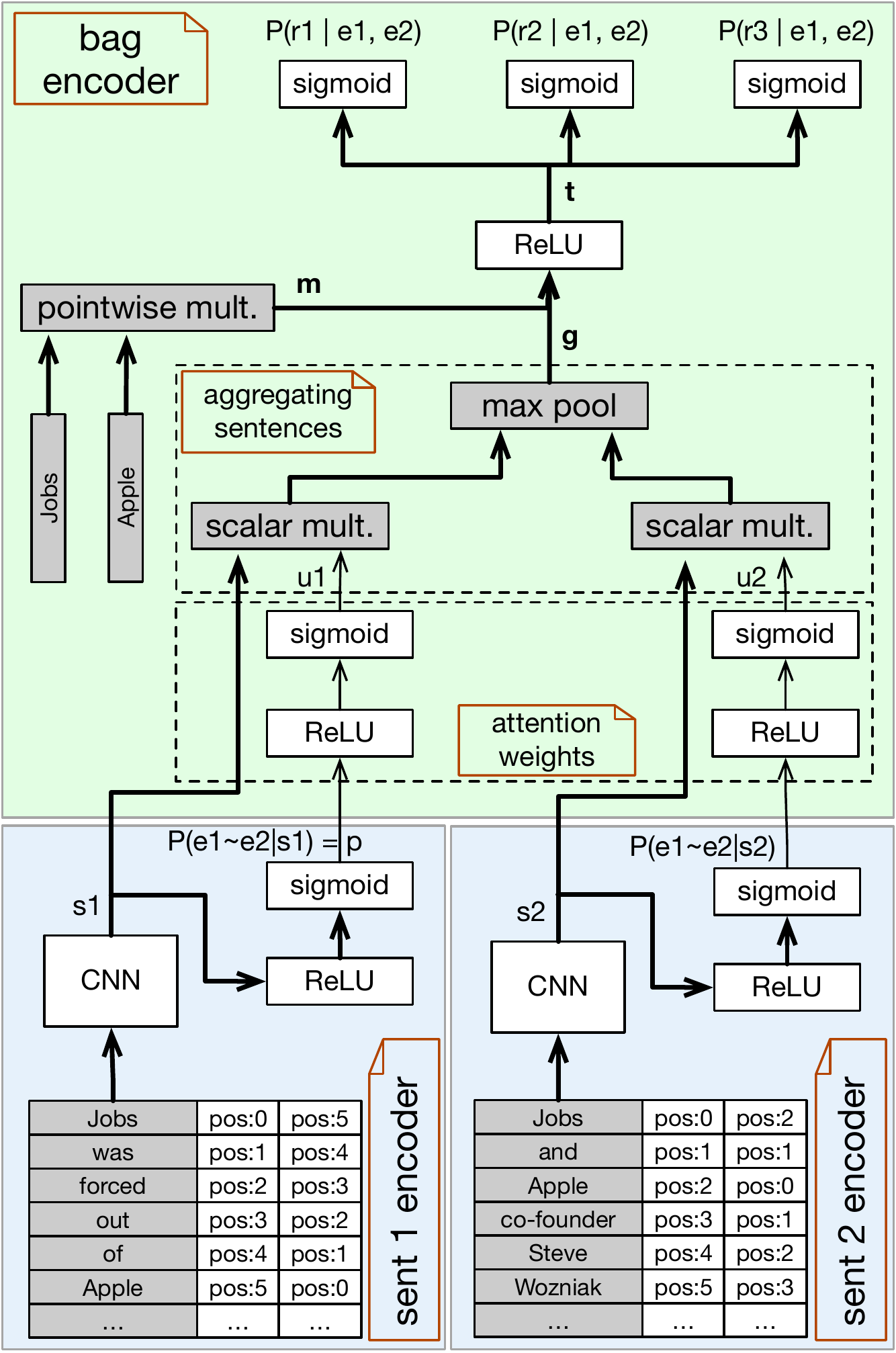}
    \vspace{-1pt}
  \caption{Blue box is the sentence encoder, 
it maps a sentence to a fixed length vector (\texttt{CNN} output)
and the probability it expresses a relation between $e_1$ and $e_2$
(\texttt{sigmoid} output).
Green box is the bag encoder, it takes encoded sentences 
and their weights and produces a fixed length vector 
(\texttt{max pool} output), concatenates it with entity
embeddings (\texttt{pointwise mult.} output)
then outputs a probability for each relation type $r$. 
White boxes contain parameters that the model learns
while gray boxes do not have learnable parameters.
Directly supervised annotations provide supervision for
$P(e_1 \sim e_2 \mid \mathbf{s})$. Distantly supervised annotations 
provide supervision for $P(r = 1 \mid e_1, e_2)$.}
    \label{fig:relex_nn}
    \vspace{-2ex}
\end{figure}

\paragraph{Input Representation.}
The input representation is illustrated graphically with a table at the bottom of Figure~\ref{fig:relex_nn}.
We map word token $i$ in the sentence $w_i$ to a pre-trained word embedding vector $\mathbf{w}_i$.\footnote{Following~\citet{lin:acl16}, we do not update the word embeddings while training the model.}
%
%
%
%
Another crucial input signal is the position of entity mentions
in each sentence $s \in B_{e_1,e_2}$.
Following \citet{zeng:coling14}, we map the distance between each word in the sentence and the entity mentions\footnote{If an entity is mentioned more than once in the sentence, we use the distance from the word to the closest entity mention. Distances greater than 30 are mapped to the embedding for distance = 30.} to a small vector of learned parameters, namely $\mathbf{d}^{e_1}_i$ and $\mathbf{d}^{e_2}_i$.



We find that adding a dropout layer with a small probability ($p=0.1$) 
before the sentence encoder reduces overfitting and improves the results. 
To summarize, the input layer for a sentence $s$ is a sequence of vectors:
\begin{align}
\mathbf{v}_i = [\mathbf{w}_i;\mathbf{d}^{e_1}_i;\mathbf{d}^{e_2}_i], \text{ for } i \in 1,\ldots,|s| \nonumber
\end{align}

\paragraph{Word Composition.}
Word composition is illustrated with the block \texttt{CNN} in the bottom part of Figure~\ref{fig:relex_nn}, which represents a convolutional neural network (CNN) with multiple filter sizes. 
The outputs of the max pool operations for different filter sizes are concatenated then
projected into a smaller vector using one feed forward linear layer. 


Sentence encoding $\mathbf{s}$ is computed as follows:
\begin{align}
\mathbf{c}_x &= \text{ CNN}_x(\mathbf{v}_1,\ldots,\mathbf{v}_{|s|}), \text{for } x \in \{2, 3, 4, 5\} \nonumber \\
\mathbf{s} &= \mathbf{W}_1 \: [\mathbf{c}_2;\mathbf{c}_3;\mathbf{c}_4;\mathbf{c}_5] + \mathbf{b}_1, \nonumber
\end{align}
where $\text{CNN}_x$ is a standard convolutional neural network with filter size $x$, $\mathbf{W}_1$ and $\mathbf{b}_1$ are model parameters
and $\mathbf{s}$ is the sentence encoding.

We feed the sentence encoding $s$ into a ReLU layer followed by a sigmoid layer with output size 1, representing $P(e_1\sim e_2\mid \mathbf{s})$, as illustrated in Figure~\ref{fig:relex_nn} (bottom):
\begin{align}
P(e_1 \sim e_2 \mid \mathbf{s}) &= \label{eq:model_binary_pred} \\
p &= \sigma(\mathbf{W}_3 \text{ReLU}(\mathbf{W}_2 \mathbf{s} + \mathbf{b}_2) + \mathbf{b}_3), \nonumber
\end{align}
where $\sigma$ is the sigmoid function and $\mathbf{W}_2, \mathbf{b}_2, \mathbf{W}_3, \mathbf{b}_3$ are model parameters.

\subsection{Bag Encoder Architecture}
\label{sec:bag_encoder_architecture}
Given a bag $B_{e_1,e_2}$ of $n \geq 1$ sentences, we compute their encodings $\mathbf{s}_1, \ldots, \mathbf{s}_n$ as described earlier and feed them into the bag encoder, which combines the information in all of the sentence encodings and predicts the probability $P(r = 1 \mid e_1, e_2), \forall r \in {\cal{R}}$.
The bag encoder also incorporates the signal $p = P(e_1 \sim e_2 \mid \mathbf{s})$ from Equation~\ref{eq:model_binary_pred} as an estimate of the degree to which sentence $s$ expresses ``some'' relation between $e_1$ and $e_2$.

\paragraph{Attention.}
To aggregate the sentence encodings
$\mathbf{s}_1, \ldots, \mathbf{s}_n$ into a fixed length vector that captures the important
features in the bag, we use attention. 
Attention has two steps: (1) computing weights for the sentences 
and (2) aggregating the weighted sentences.
Weights can be uniform, or computed using a sigmoid or softmax. 
Weighted sentences can be aggregated using average pooling or max pooling.
Prior work \cite{jiang:coling16,lin:acl16,ji:acl17} have explored
some of these combinations but not all of them. 
In the ablation experiments, 
we try all 
combinations and we find that the (sigmoid, max pooling) attention
gives the best result. We discuss the intuition behind this in Section~\ref{sec:ablation}. 
For the rest of this section, we will explain the architecture of our network
assuming a (sigmoid, max pooling) attention.

Given the encoding $\mathbf{s}_j$ and an unnormalized weight $u_j$ for each sentence $s_j \in B_{e_1,e_2}$,
the bag encoding $\mathbf{g}$ is a vector with the same dimensionality as $\mathbf{s}_j$. 
With (sigmoid, max pooling) attention, each sentence vector is multiplied 
by the corresponding weight, then we do a ``dimension-wise" max pooling (taking the maximum of each dimension across all sentences, 
not the other way around). 
The $k$-th element of the bag encoding $\mathbf{g}$  is computed as:
\begin{equation}
\label{eq:maxpool_attention}
\mathbf{g}_j[k] = \text{max}_{j\in 1, \ldots, n} \{ \mathbf{s}_j[k] \times \sigma(u_j) \}. \nonumber
\end{equation}

As shown in Figure~\ref{fig:relex_nn}, we do not directly use the $p$ from Equation~\ref{eq:model_binary_pred} as attention weights. Instead, 
we found it useful to feed it into more non-linearities.
The unnormalized attention weight for $s_j$ is computed as: 
\begin{equation*}
u_j = \textbf{W}_7 \: \text{ReLU}(\mathbf{W}_6 \: p + \textbf{b}_6) + \textbf{b}_7.
\end{equation*}

\paragraph{Entity Embeddings.}
\label{sec:e_embd}
Following \citet{ji:aaai17}, we use entity embeddings to improve our model of relations in the distant supervision setting, although 
our formulation is closer to that of \citet{yang:iclr15} who used point-wise multiplication of
entity embeddings: ${\mathbf{m} = \mathbf{e}_1 \odot \mathbf{e}_2},$
where $\odot$ is point-wise multiplication, and $\mathbf{e}_1$ and $\mathbf{e}_2$ are the embeddings of $e_1$ and $e_2$, respectively.
In order to improve the coverage of entity embeddings, we use pretrained GloVe vectors ~\cite{pennington:emnlp14} (same embeddings used in the input layer).
For entities with multiple words, like ``Steve Jobs'', the vector for the entity is the average of the GloVe vectors of its individual words. 
If the entity is expressed differently across sentences, 
we average the vectors of the different mentions. 
As discussed in Section~\ref{sec:detailed_results}, this leads 
to big improvement in the results, and we believe there is still big room 
for improvement from having better representation for entities.
We feed the output $\mathbf{m}$ as additional input to the last block of our model. 

\paragraph{Output Layer.}
The final step is to use the bag encoding $\mathbf{g}$ and the entity pair encoding $\mathbf{m}$ to predict a set of
relations $L^{\text{pred}}$ which is a standard multilabel classification problem. 
We concatenate $\mathbf{g}$ and $\mathbf{m}$ and feed them into a feedforward layer with ReLU non-linearity, followed by a sigmoid layer with an output size of $|{\cal{R}}|$:
\begin{align}
\mathbf{t} &= \text{ReLU}(\mathbf{W}_4 [\mathbf{g};\mathbf{m}] + \mathbf{b}_4)  \nonumber \\
P(\mathbf{r} = 1 \mid e_1, e_2) &= \sigma( \mathbf{W}_5 \mathbf{t} + \mathbf{b}_5), \nonumber
\end{align}
where $\mathbf{r}$ is a vector of Bernoulli variables each of which corresponds to one of the relations in $\cal{R}$. 
This is the final output of the model.

\subsection{Model Training}
To train the model on the distant supervision data, 
we use binary cross-entropy loss between the model predictions and the labels obtained with distant supervision, i.e.,
\begin{align}
\text{DistSupLoss} = \sum_{B_{e_1,e_2}} - \log P(\mathbf{r} = \mathbf{r}^{\text{distant}} \mid e_1, e_2) \nonumber
\end{align}
where $\mathbf{r}^{\text{distant}}[k] = 1$ indicates that the tuple $(e_1, r_k, e_2)$ is in the knowledge base.

\label{sec:sent_sup}
In addition to the distant supervision, we want to improve the results
by incorporating an additional direct supervision.
A straightforward way to combine them is to create
singleton bags for direct supervision labels, and add the bags to those obtained with distant supervision.
However, results in Section~\ref{sec:ablation}
show that this approach does not improve the results. 
Instead, a better use of the direct supervision is to improve the model's ability to predict  
the potential usefulness of a sentence.
According to our analysis of baseline models, distinguishing between positive and negative examples is the real bottleneck in the task. 
Therefore, we use the direct supervision data to supervise
$P(e_1 \sim e_2 \mid \mathbf{s})$. 
This supervision serves two purposes: it improves our encoding of each sentence, and improves the weights used by the attention to decide which sentences should contribute more to the bag encoding.
It also has the side benefit of not requiring the same set of relation
types as that of the distant supervision data, because we only 
care about if there exists \emph{some relevant relation} or not between the entities.

We minimize log loss of gold labels in the direct supervision data ${\cal{D}}$ according to the model described in Equation~\ref{eq:model_binary_pred}:
\begin{align}
\text{DirectSupLoss} = \sum_{s,l^{\text{gold}} \in {\cal{D}}} - \log P(l = l^{\text{gold}}\mid \mathbf{s}) \nonumber
\end{align}
where ${\cal{D}}$ is all the direct supervision data
and all distantly-supervised \emph{negative} examples.\footnote{We note that the distantly supervised negative examples may still be noisy. However, given that relations tend to be sparse, the signal to noise is high.}


We jointly train the model on both types of supervision. 
The model loss is a weighted sum of the direct supervision
and distant supervision losses, 
\begin{equation}
\label{eq:lambda}
\text{loss} =  \dfrac{1}{\lambda + 1} \text{DistSupLoss}
+ \dfrac{\lambda}{\lambda + 1} \text{DirectSupLoss}
\end{equation}
where $\lambda$ is a parameter that controls the contribution of each
loss, tuned on a validation set. 

\section{Experiments}

\subsection{Data and Setup} 
\label{sec:setup}
This section discusses datasets, metrics, configurations and 
the models we are comparing with. 

\paragraph{Distant Supervision Dataset (DistSup).}
The FB-NYT dataset\footnote{\url{http://iesl.cs.umass.edu/riedel/ecml/}}
introduced in \citet{riedel:ecml10} was generated 
by aligning Freebase facts with New York Times articles.
The dataset has 52 relations with the most common being ``location'', ``nationality'', ``capital'', ``place\_lived'' and ``neighborhood\_of''. 
They used the articles of \textit{2005} and \textit{2006}
for training, and \textit{2007} for testing.
Recent prior work~\citep{lin:acl16,liu:emnlp17,huang:emnlp17}
changed the original dataset. They used all articles for training except those from \textit{2007}, which they left for testing as in \citet{riedel:ecml10}.
We use the modified dataset which was made available 
by \citet{lin:acl16}.\footnote{\url{https://github.com/thunlp/NRE}} 
The table below shows the dataset size.
\begin{center}
\small
\setlength{\tabcolsep}{5pt}
\begin{tabular}{@{}rrr@{}}
\toprule
      & Train & Test\\
\midrule
Positive bags & 16,625 & 1,950\\
Negative bags & 236,811 & 94,917  \\
Sentences & 472,963 & 172,448 \\
 \bottomrule
 \end{tabular}
\end{center}

\paragraph{Direct Supervision Dataset (DirectSup).}
Our direct supervision dataset was made available by \citet{angeli:emnlp14} and it was collected in an active learning framework.
The dataset consists of sentences annotated with entities and 
their relations. It has 22,766 positive examples
for 41 relation types in addition to 11,049 negative examples.
To use this dataset as supervision for $P(e_1 \sim e_2 \mid \mathbf{s})$, 
we replace the relation types of positive examples with $1$s 
and label negative examples with $0$s.


\paragraph{Metrics.} 
Prior work used precision-recall (PR) curves to show results
on the FB-NYT dataset. 
In this multilabel classification setting, the PR curve is constructed using
the model predictions on all entity pairs in the test set for all relation types sorted by the confidence scores
from highest to lowest.
Different thresholds correspond to different points on the PR curve. 
We use the area under the PR curve (AUC) for early stopping and hyperparameter tuning.
Following previous work on this dataset, we only keep points on the PR curve with recall below $0.4$, focusing on the high-precision low-recall part of the PR curve.
As a result, the largest possible value for AUC is $0.4$.

\paragraph{Configurations.}
The FB-NYT dataset does not have a validation set for hyperparameter tuning
and early stopping. \newcite{liu:emnlp17} use the test set for validation,
\newcite{lin:acl16} use 3-fold cross validation, 
and \newcite{vashishth:emnlp18} split the training set into 80\% training 
and 20\% testing. In our experiments, we use 90\% of the training set for training and keep the other 10\% for validation.
The main hyperparameter we tune is lambda (section~\ref{sec:lambda}).

The pre-trained word embeddings we use are 300-dimensional GloVe vectors, trained on 42B tokens.
Since we do not update word embeddings while training the model, we define our vocabulary as any word which appears in the training, validation or test sets with frequency greater than two. 
When a word with a hyphen (e.g., `five-star') is not in the GloVe vocabulary, we average the embeddings of its subcomponents.
Otherwise, all OOV words are assigned the same random vector (normal with mean 0 and standard deviation 0.05).

Our model is implemented using PyTorch and AllenNLP~\citep{gardner:arxiv17}
and trained on machines with P100 GPUs. Each run takes five hours on average. 
We train for a maximum of 50 epoch, and use early stopping with patience $= 3$. 
Each dataset is split into minibatches of size 32 and randomly shuffled before every epoch. 
We use the Adam optimizer with its default PyTorch parameters.
We run every configuration with three random seeds and report the PR curve for the run with
the best validation AUC. 
In the controlled experiments, we report the mean and standard deviation of the AUC across runs.


\paragraph{Compared Models.}
Our best model (Section~\ref{sec:model}) is trained on
the DistSup and DirectSup datasets in our multitask 
setup and it uses (sigmoid, max pooling) attention.
\textbf{Baseline} is the same model described in Section~\ref{sec:model}
but trained only on the DistSup
dataset and uses the more common (softmax, average pooling) attention.
This baseline is our implementation of the
\textbf{PCNN+ATT} model~\cite{lin:acl16}
with two main differences; they use piecewise convolutional neural networks ~\cite[PCNNs][]{pennington:emnlp14} instead of
CNNs, and we add entity
embeddings before the output layer.\footnote{Contrary to the results in~\citet{pennington:emnlp14}, we found CNNs to give better results than PCNNs in our experiments. \newcite{lin:acl16} also compute unnormalized attention weights as ${o_j = \mathbf{s}_j \times \mathbf{A} \times \mathbf{q}}$ where $\mathbf{s}_j$ is the sentence encoding, $\mathbf{A}$ is a diagonal matrix and  $\mathbf{q}$ is the query vector. In our experiments, we found that implementing it as a feedforward layer with output size $= 1$ works better. All our results  use the feedforward implementation.} 
We also compare our results to the state of the art model \textbf{RESIDE}~\citep{vashishth:emnlp18},
which uses graph convolution over dependency parse trees,
OpenIE extractions and entity type constraints. 

\subsection{Main Results}
\label{sec:detailed_results}
\begin{figure}[t]
    \centering
    \begin{subfigure}{0.40\textwidth}
        \centering
        \includegraphics[width=\columnwidth]{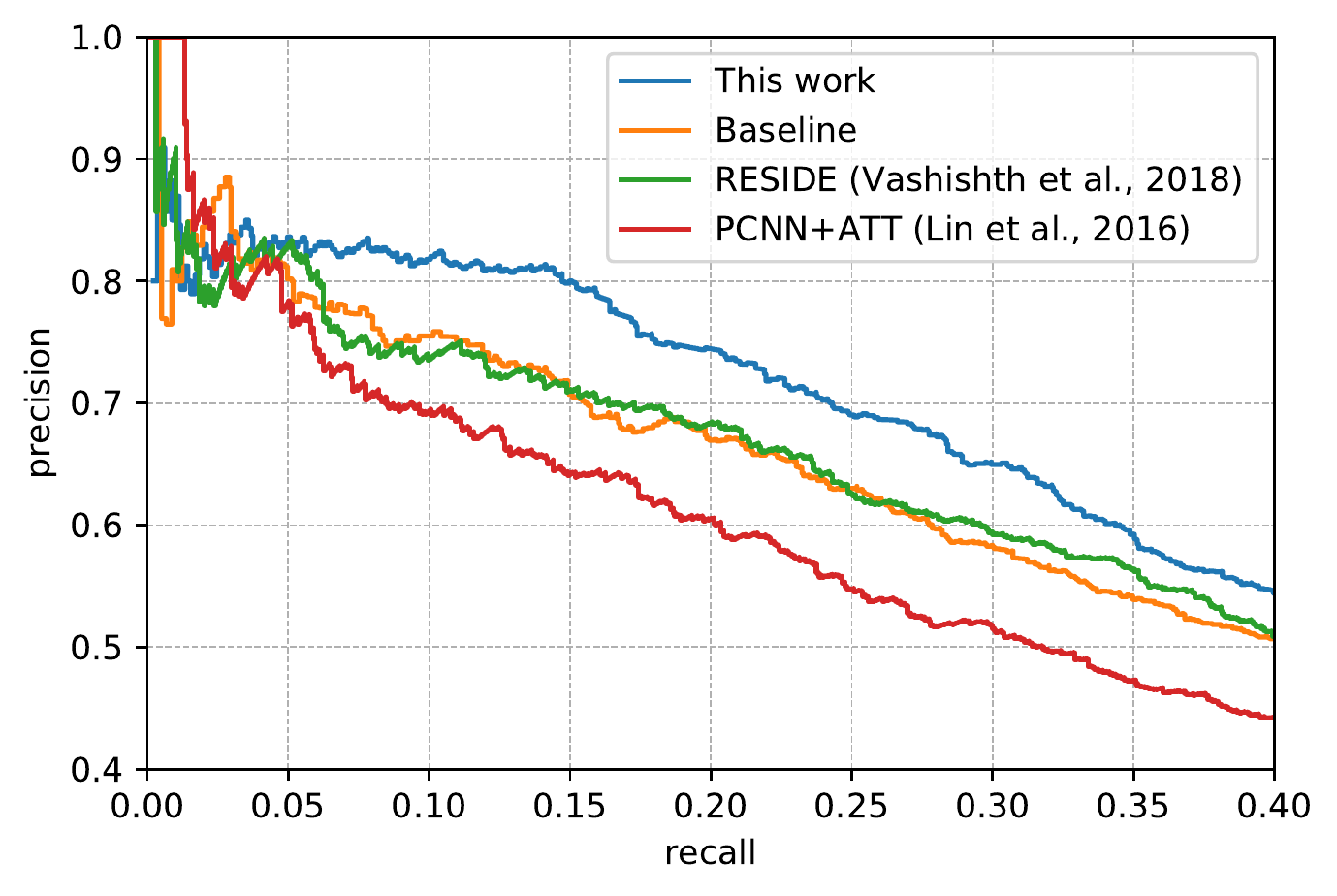}
    \end{subfigure}%
    \\
    \begin{subfigure}{0.45\textwidth}
		\centering
        \small
\setlength{\tabcolsep}{5pt}
\begin{tabular}{@{}rl@{}}
\toprule
 \qquad \quad \textbf{Model} & \textbf{AUC} \\ 
\midrule
PCNN+ATT~\citep{lin:acl16} & 0.247 \\
RESIDE~\citep{vashishth:emnlp18} & 0.271 \\
Our baseline & 0.272$_{\pm 0.005}$ \\
\hline
This work & \textbf{0.283$_{\pm 0.007}$} \\
\bottomrule
\end{tabular}
    \end{subfigure}
    \caption{\label{fig:results}Precision-recall curves and their AUC comparing
    our model with the baseline and existing models.
    Baseline is trained on DistSup and uses (softmax, average pooling) attention.
    Our best model is trained using multitask learning on DistSup and DirectSup and uses 
    (sigmoid, max pooling) attention.
    Results of~\citet{lin:acl16} and~\citet{vashishth:emnlp18} are copied from 
their papers.}
\end{figure}
Figure~\ref{fig:results} summarizes
the main results of our experiments.
First, we note that ``our baseline'' outperforms PCNN+ATT~\citep{lin:acl16} despite 
using the same training data (DistSup) and the same form of attention
(softmax, average pooling), which confirms that we are building on a strong baseline.
The improved results in our baseline are due to using CNNs instead of PCNNs, and using entity embeddings. 

\paragraph{A new state-of-the-art.}
Adding DirectSup in our multitask learning setup and using
(sigmoid, max pooling) attention gives us the best result, outperforming
our baseline that doesn't use either by 4.4\% relative AUC increase,
and achieves a new state-of-the-art result outperforming~\cite{{vashishth:emnlp18}}.

We note that the improved results reported here conflate additional supervision and model improvements.
Next, we report the results of controlled experiments to tease apart the contributions of different components.
%


\begin{table}
\centering
\scriptsize
\setlength{\tabcolsep}{4pt}
\begin{tabular}{@{}p{20pt}llll@{}}
\toprule
pooling type & supervision signal  & \multicolumn{3}{c}{attention weight computation} \\
\cmidrule(lr){3-5}
   &  & \multicolumn{1}{c}{uniform}   & \multicolumn{1}{c}{softmax} & \multicolumn{1}{c}{sigmoid}  \\
\midrule
\multirow{3}{1pt}{average pooling}
& DistSup & 0.244\textsubscript{$\pm$ 0.008}  & 0.272\textsubscript{$\pm$ 0.005} & 0.258\textsubscript{$\pm$ 0.020} \\
& DistSup + DirectSup & 0.224\textsubscript{$\pm$ 0.009} & 0.272\textsubscript{$\pm$ 0.009} & 0.256\textsubscript{$\pm$ 0.009}\\
& MultiTask (our model) & 0.220\textsubscript{$\pm$ 0.012} & 0.262\textsubscript{$\pm$ 0.014} & 0.258\textsubscript{$\pm$ 0.015}\\
\hline
\multirow{3}{1pt}{max pooling}
& DistSup & 0.277\textsubscript{$\pm$ 0.009} & 0.278\textsubscript{$\pm$ 0.001} & 0.274\textsubscript{$\pm$ 0.004} \\
& DistSup + DirectSup  & 0.269\textsubscript{$\pm$ 0.003} & 0.269\textsubscript{$\pm$ 0.005} & 0.277\textsubscript{$\pm$ 0.012} \\
& MultiTask (our model) & 0.266\textsubscript{$\pm$ 0.007} & 0.280\textsubscript{$\pm$ 0.004} &0.283\textsubscript{$\pm$ 0.007} \\
\bottomrule
\end{tabular}
\caption{Controlled experiments for
a) how the supervised data is used in the model, 
b) which function is used to compute attention weights,
c) how sentence embeddings are aggregated}
\label{tab:ablation}
\vspace{-2ex}
\end{table}

Table~\ref{tab:ablation} summarizes results of our controlled experiments 
showing the impact of how the training data is used, 
and the impact of different configurations of the attention (computing weights
and aggregating vectors).
The model can be trained on DistSup only, DistSup + DirectSup together
as one dataset with DirectSup expressed as singleton bags, or DistSup + DirectSup in our multitask setup.
Attention weights can be 
uniform, 
or computed using softmax or sigmoid.\footnote{We also tried normalizing sigmoid weights as suggested in~\citep{rei:naacl18},
but this did not work better than regular sigmoid or softmax.}
Sentence vectors
are aggregated by weighting them then averaging (average pooling)
or weighting them then taking the max of each dimension (max pooling). 
(Uniform weights, average pooling) and (softmax, average pooling) were used by~\newcite{lin:acl16}, (sigmoid, average pooling) was proposed by~\newcite{ji:acl17}
but for a different task, and (uniform weights, max pooling) is used by~\newcite{jiang:coling16}. To the best of our knowledge, (softmax, max pooling) and (sigmoid, max pooling) have not been explored before.

\paragraph{Pooling type.}
\label{sec:ablation}
Results in Table~\ref{tab:ablation} show that aggregating sentence vectors using max pooling
generally works better than average pooling. This is because max pooling might be better at picking out
useful features (dimensions) from each sentence.

\paragraph{Supervision signal.}
The second dimension of comparison is the use of the supervision signal used to train the model. The table shows 
that training on DistSup + DirectSup, where the DirectSup dataset is simply used 
as additional bags, can hurt the performance. 
We hypothesize that this is because the DirectSup data change the distribution of relation types in the training set from the test set.
However, using DirectSup as supervision for the attention weights in our 
multitask learning setup leads to considerable improvements
(1\% and 3\% relative AUC increase using softmax and sigmoid respectively)
because it leads to better attention weights
and improves the model's ability to filter noisy sentences.

\paragraph{Attention weight computation.} Finally, comparing uniform weights, softmax and sigmoid.
We found the result to depend on the available level of supervision.
With DistSup only,
the results of all three are comparable with softmax being slightly better.
However, when we have good attention weights (as provided by the multitask learning),
softmax and sigmoid work better than uniform weights where sigmoid gives the best result with 6\% relative AUC increase.
Sigmoid works better than softmax, because softmax assumes that exactly one sentence is correct by forcing the probabilities to sum to 1. This assumption is not correct for this task, because zero or many sentences could potentially be relevant.
On the other hand, sigmoidal attention weights does not make this assumption, which 
gives rise to more informative attention weights in cases where all sentences are not useful, 
or when multiple ones are. This makes the sigmoidal attention weights a better modeling for the problem (assuming reliable attention weights).

\begin{figure}[t]
    \centering
    \vspace{-3ex}
    \includegraphics[width=0.8\columnwidth]{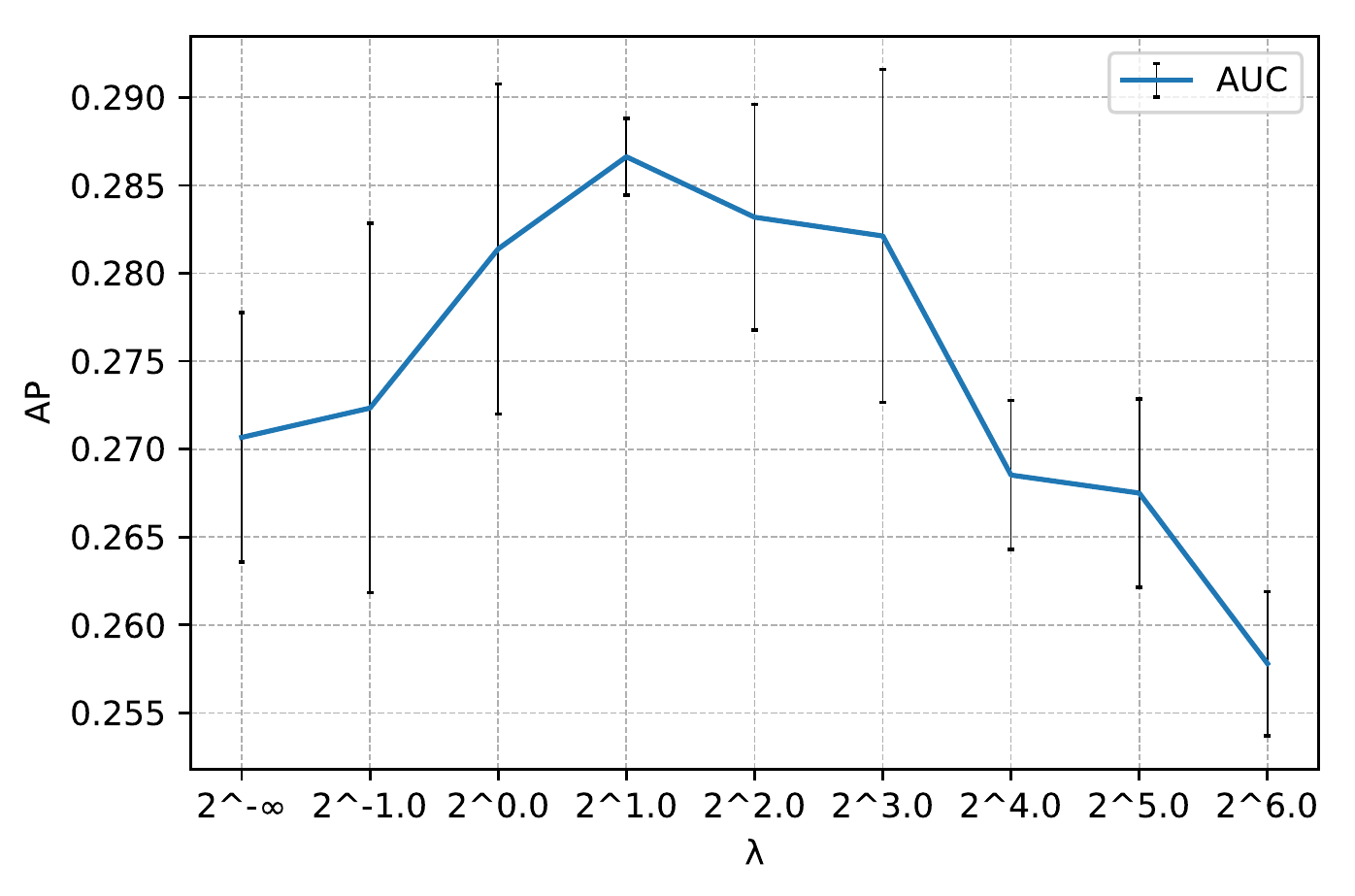}
    \vspace{-0.4cm}
  \caption{AUC at different $\lambda$. X-axis is in log-scale.}
    \label{fig:lambda}
\end{figure} 
\subsection{Selecting Lambda.}
\label{sec:lambda}
Although we did not spend much time tuning hyperparameters, we made sure to carefully tune $\lambda$ (Equation \ref{eq:lambda})
which balances the contribution of the two losses of the multitask learning.
Early experiments showed that DirectSupLoss is typically smaller than DistSupLoss,
so we experimented with $\lambda \in \{0, 0.5, 1, 2, 4, 8, 16, 32, 64\}$.
Figure~\ref{fig:lambda} shows AUC results for different values
of $\lambda$, where each point is the average of three runs. 
It is clear that picking the right value for $\lambda$ has a big 
impact on the final result.

\subsection{Qualitative Analysis.}
\begin{table*}[t]
\centering
\scriptsize
\setlength{\tabcolsep}{3pt}
\begin{tabular}{@{}ccp{0.85\textwidth}@{}}
\toprule
Baseline & This work & \multicolumn{1}{c}{Sentences}\\ 
\midrule
 0.00 & 0.029 & You can line up along the route to cheer for the 32,000 riders, whose 42-mile trip will start in battery park and end with a festival at \textbf{Fort Wadsworth} on \textbf{Staten Island} . \\ \hline
0.00 & 0.026 & Gateway is a home to the nation's oldest continuing operating lighthouse, Sandy Hook lighthouse, built in 1764; Floyd Bennett field in Brooklyn, which was the city's first municipal airfield; \textbf{Fort Wadsworth} on \textbf{Staten Island}, which predates the revolutionary war. \\ \hline
0.99  & 0.027 & home energy smart fair, gateway national recreation area, \textbf{Fort Wadsworth} visitor center, bay street and school road, \textbf{Staten Island}. \\ 
\bottomrule
 \end{tabular}
 \caption{Weights assigned to sentences by 
 our baseline and our best model. The baseline incorrectly predicts ``no\_relation'', while our best model correctly predicts ``neighbourhood\_of'' for this bag.}
  \label{tab:cp}
  \vspace{-3ex}
\end{table*}

An example of a positive bag is shown in Table~\ref{tab:cp}.
Our best model (Multitask, sigmoid, max pooling) assigns the most weight to the first sentence while the baseline (DistSup, softmax, average pooling) assigns the most weight to the last sentence (which is less informative for the relation between the two entities).
Also, the baseline does not use the other two sentences because their weights are dominated by the last one.

\section{Related Work}
\label{sec:related}

\paragraph{Distant Supervision.}
The term `distant supervision' was coined by \newcite{mintz:acl09} who used relation instances in a KB 
to identify sentences in a text corpus where two related entities are mentioned, then developed a classifier to predict the relation.
Researchers have since extended this approach further~\cite[e.g.,][]{takamatsu:acl12,min:naacl13,riedel:naacl13,koch:emnlp14}. 

A key source of noise in distant supervision is that sentences may mention two related entities without expressing the relation between them.
\newcite{hoffmann:acl11} used multi-instance learning to address this problem by developing a graphical model for each entity pair which includes a latent variable for each sentence to explicitly indicate the relation expressed by that sentence, if any.
Our model can be viewed as an extension of~\newcite{hoffmann:acl11} where the sentence-bound latent variables can also be directly supervised in some of the training examples.



\paragraph{Neural Models for Distant Supervision.}
More recently, neural models have been effectively used to model textual relations~\cite[e.g.,][]{hashimoto:emnlp13,zeng:coling14,nguyen:naacl2015}.
Focusing on distantly supervised models, \newcite{zeng:emnlp15} proposed a neural implementation of multi-instance learning to leverage multiple sentences which mention an entity pair in distantly supervised relation extraction.
However, their model
picks only one sentence to represent an entity pair,
which wastes the information in the neglected sentences.
\newcite{jiang:coling16} addresses this limitation by max pooling the vector encodings of all input sentences for a given entity pair. 
\newcite{lin:acl16} independently proposed to use attention to address the same limitation, and 
\newcite{du:emnlp18} improved by using multilevel self-attention.
To account for the noise in distant supervision labels, \newcite{liu:emnlp17,luo:acl17,wang:emnlp18} suggested different ways of using ``soft labels'' that do not necessarily agree with the distant supervision labels.
\newcite{ye:acl17} proposed a method for leveraging dependencies between different relations in a pairwise ranking framework, while
\newcite{han:emnlp18} arranged the relation types in a hierarchy aiming for better generalization for relations that do not have enough training data.
To improve using additional resources, 
\newcite{vashishth:emnlp18} 
used graph convolution over dependency parse,
OpenIE extractions and entity type constraints, 
and 
\newcite{liu:emnlp18} used parse trees to prune irrelevant information from the sentences.

\paragraph{Combining Direct and Distant Supervision.}

Despite the substantial amount of work on both directly and distantly supervised relation extraction, the question of how to combine both signals has not received 
the same attention. 
~\citet{pershina:acl14} trained MIML-RE from ~\citep{surdeanu:emnlp12} on both types of supervision by locking the latent variables on the sentences to the supervised labels.
\citet{angeli:emnlp14} and \citet{liu:naacl16} presented active learning models that select sentences to annotate and incorporate in the same manner.
~\citet{pershina:acl14} and \citet{liu:naacl16} also tried simple baseline of including the labeled sentences as singleton bags. 
~\citet{pershina:acl14} did not find this beneficial, which agrees with our results in Section~\ref{sec:ablation}, while \citet{liu:naacl16} found the addition of singleton bags to work well.


Our work is addressing the same problem, but combining both signals in a state-of-the-art neural network model, and we do not require the two datasets to have the same set of relation types.



\section{Conclusion}
We improve neural network models for relation extraction 
by combining distant and direct supervision data.
Our network uses attention to attend to relevant sentences, and we use the direct supervision to improve attention weights, thus improving the model's ability to find sentences that are likely to express a relation.
We also found that sigmoidal attention weights with max pooling achieves better performance than the commonly used weighted average attention.
Our model combining both forms of supervision achieves a new state-of-the-art result on the FB-NYT dataset with a 4.4\% relative AUC increase
than our baseline without the additional supervision.


\section*{Acknowledgments}
All experiments were performed on \url{beaker.org}. Computations on \url{beaker.org} were supported in part by credits from Google Cloud.
\bibliography{citations}
\bibliographystyle{acl_natbib}

\end{document}